\PassOptionsToPackage{table,xcdraw}{xcolor}

\documentclass{article} 
\usepackage{preprint,times}

\usepackage{xcolor}

\usepackage{amsmath,amsfonts,bm}

\def\eqref#1{equation~\ref{#1}}

\def\1{\bm{1}}

\DeclareMathAlphabet{\mathsfit}{\encodingdefault}{\sfdefault}{m}{sl}
\SetMathAlphabet{\mathsfit}{bold}{\encodingdefault}{\sfdefault}{bx}{n}

\usepackage{amsmath,amssymb}
\usepackage{mathtools}
\usepackage{bbm}
\usepackage{dsfont}

\usepackage{graphicx}
\usepackage[inkscape]{svg}
\graphicspath{{figures/}}

\usepackage{array}
\usepackage{booktabs}
\usepackage{multirow}
\usepackage{tabularx}
\usepackage{colortbl}
\usepackage{ragged2e}

\usepackage{caption}
\usepackage{subcaption}

\usepackage{microtype}
\usepackage{xspace}
\usepackage{chngcntr}
\usepackage{enumerate}
\usepackage{enumitem}
\usepackage[normalem]{ulem}
\usepackage{soul}
\usepackage{mdframed}
\usepackage{listings}

\usepackage{natbib}
\usepackage{url}
\usepackage{hyperref}
\usepackage{tablefootnote}

\definecolor{citeblue}{RGB}{35,131,75}
\hypersetup{
  colorlinks=true,
  citecolor=citeblue
}

\title{Your Unlearning Gives You Away: Identifying Erased Concepts in Diffusion Models}

\author{\parbox[t]{\textwidth}{\centering
Kaiyuan Deng$^{1}$\quad Yuchen Li$^{2}$\quad Yang Xiao$^{3}$\quad Bo Hui$^{3}$\quad Geng Yuan$^{4}$\quad Xiaolong Ma$^{1}$ \\
\normalfont $^1$The University of Arizona\quad $^2$Peking University \\
\normalfont $^3$The University of Tulsa\quad $^4$University of Georgia \\
\normalfont \texttt{kaiyuan0415@arizona.edu}}
}

\iclrfinalcopy

\begin{document}

\maketitle
\vspace{-10pt}


\begin{abstract}
\label{abs}

Existing attacks on unlearned diffusion models assume that the erased concepts are known in advance and focus on recovering them. In practice, however, model providers may not disclose which concepts have been removed, and even with access to the original base model, an adversary may still lack a clear target to attack. In this paper, we aim to answer the following critical but overlooked questions: \emph{\textbf{which concepts have been erased from the model, and how many have been erased in total?}} 
To this end, we present \textsc{Tracer}, a framework that rapidly and accurately identifies erased concepts and estimates their number.
\textsc{Tracer} efficiently identifies erased concepts without generating and classifying images. By combining lightweight spectral analysis of weight footprints, it enables efficient search over large candidate vocabularies.
To distinguish multiple erased concepts, we introduce a footprint coverage objective that guides sequential discovery. \textsc{Tracer} estimates the number of erased concepts by detecting a sharp decline in candidate confidence as the selected concepts account for the erasure footprint, without requiring labeled examples for calibration. The framework requires only lightweight linear algebra and limited forward probes, with no prior knowledge of the unlearning algorithm. 
Experiments across text-to-image and text-to-video backbones and diverse unlearning methods demonstrate that \textsc{Tracer} identifies erased concepts and estimates their number \textit{in seconds}, achieving $\boldsymbol{150}$--$\boldsymbol{137{,}000\times}$ and $\boldsymbol{133}$--$\boldsymbol{20{,}000\times}$ speedups over MIA and brute-force search on image and video models, respectively, with substantially higher identification accuracy.

\end{abstract}

\vspace{-8pt}
\section{Introduction}
\label{intro}

Diffusion models~\citep{rombach2022high, podell2023sdxl, ding2022cogview2} exhibit powerful generative capabilities.
Despite these advances, their large-scale deployment has raised concerns about copyright infringement and harmful or biased content generation.
Concept unlearning~\citep{bourtoule2021machine, xu2024machine} aims to suppress designated concepts while preserving other generative capabilities.
Most existing methods achieve this by modifying pretrained model weights, demonstrating promising results in unlearning individual concepts or sets of concepts.


Existing attacks exploit vulnerabilities in unlearned models to regenerate erased content, but they typically assume prior knowledge of the specific concepts to recover~\citep{lu2026concepts, pham2023circumventing, rusanovsky2025memories, zhang2024generate, chin2023prompting4debugging, tsai2024ring, yang2024mma, petsiuk2024concept, liu2026image, xiang2026tina, horwitz2024recovering}. In practice, model providers may not disclose what they have removed.
Even access to the original base model does not directly reveal which of its many concepts the provider chose to erase. For example, a provider may erase a small set of individuals' identities for privacy reasons. Although the base model may still generate depictions of these individuals, access to this capability alone does not tell an adversary who was selected for protection. Identifying the erased concepts therefore provides information beyond access to the original model: it reveals the provider's undisclosed targets and allows an adversary to focus subsequent attacks on the content the provider sought to suppress. This raises two critical but overlooked questions: \emph{\textbf{without any prior knowledge of the erasure targets, can an adversary know which concepts have been erased from the model, and how many have been erased in total?}} To the best of our knowledge, no existing method systematically addresses this identification problem in unlearned diffusion models.

\begin{figure}[t]
\centering
\includegraphics[width=\linewidth]{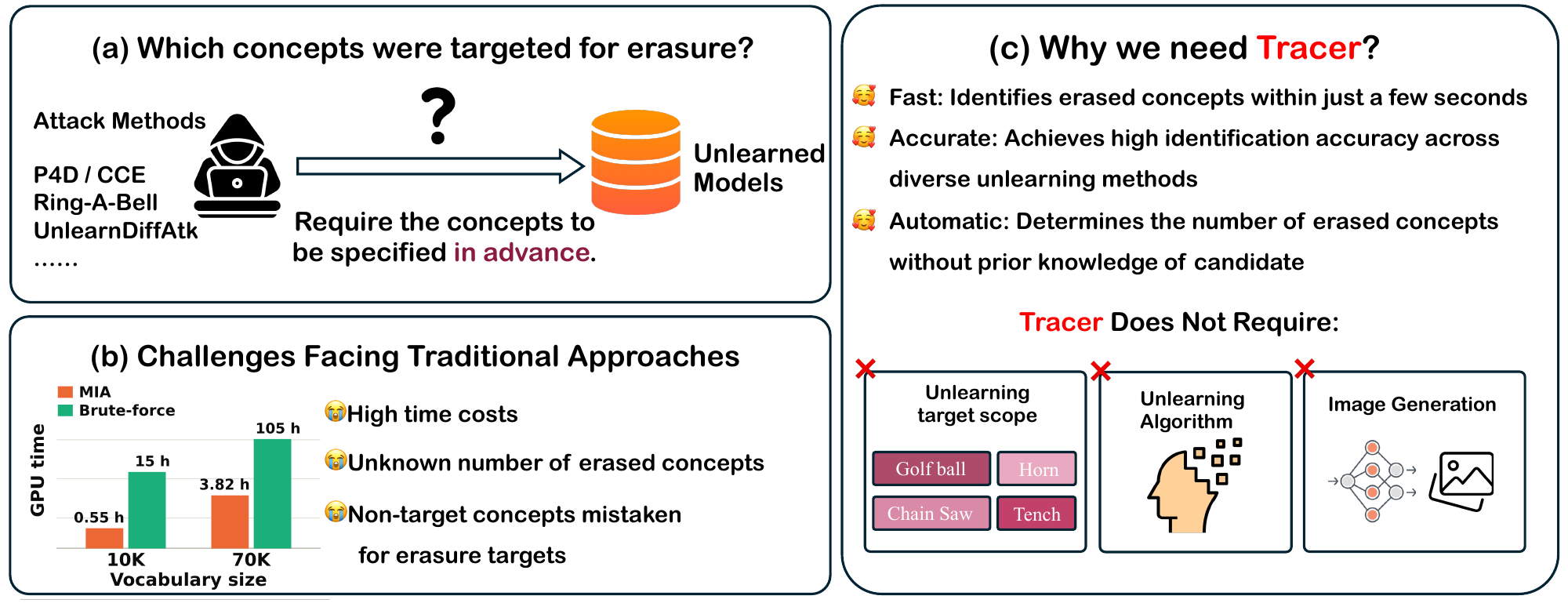}
\caption{(a) Motivation: existing unlearning attacks assume
known unlearning targets, leaving the identification of
targets unexplored.
(b) Current problem: adapted MIA and brute-force method
incur high computational costs and struggle to determine
target counts and distinguish targets from affected non-targets.
(c) \textsc{Tracer}: identifies erased concepts and estimates
their number in seconds without prior target knowledge,
knowledge of the unlearning algorithm, or image generation.}
\label{fig:teaser}
\vspace{-14pt}
\end{figure}

Potential approaches include adapting traditional membership inference attacks (MIAs)~\citep{fu2023probabilistic, kong2024efficient, duan2023diffusion, wu2022membership, matsumoto2023membership} or comparing images generated by the original and unlearned models for each candidate.
However, extending these approaches to reliable and efficient identification of unknown erasure targets presents challenges.
\emph{\textbf{First}}, unlearning can affect both target and non-target concepts.
Changes in membership scores or generated content therefore cannot reliably distinguish directly targeted concepts from indirectly affected ones.
\emph{\textbf{Second}}, the number of erased concepts is unknown.
Identifying the target set requires determining which candidates to select and when to stop, without target labels for calibration.
\emph{\textbf{Third}}, both approaches incur substantial computational costs even with a predefined candidate set.
Without a narrowly specified set, the search may span a vocabulary containing tens of thousands of words, making repeated generation or MIA-based probing prohibitively expensive.


In this paper, we introduce a novel method, \underline{Tra}cing \underline{C}oncepts \underline{Er}ased by Unlearning (\textbf{\textsc{Tracer}}), for rapidly and accurately identifying erased concepts and estimates their number \textbf{\textit{without any prior knowledge}} of the erasure targets. An intuitive but inconclusive approach is to compare the unlearned model weights with its original base model. However, these weight differences span distinct representation spaces, making it difficult to map them directly to erased concepts. \textit{\ul{To turn these differences into usable evidence}}, \textsc{Tracer} constructs Gram matrices from weight differences and uses a small set of probe words to identify compatible representation groups, then applies spectral reweighting to balance signals before aggregation, producing an erasure footprint for each group. These footprints provide the basis for identifying candidate concepts whose representations collectively explain the observed changes. 
\textit{\ul{To identify multiple erased concepts}}, we formulate identification as a subspace coverage problem, which measures how much of the erasure footprint a set of candidate concepts can collectively explain. Specifically, starting from an empty set, \textsc{Tracer} iteratively adds the candidate that best explains the remaining footprint energy, reducing redundant selections and directing the search toward other erased concepts. \textit{\ul{To estimate the number of erased concepts}}, we track candidate confidence throughout this coverage-based selection process. As the selected concepts account for the dominant footprint energy, a sharp decline in confidence indicates a cutoff in the selected sequence, allowing us to estimate the erased-set size without labeled calibration data. We summarize our contributions as follows:
\begin{itemize}[leftmargin=1.5em, itemsep=1pt, topsep=1.5pt] 
\item To the best of our knowledge, we are the \textit{first} to systematically study erasure-target identification in unlearned diffusion models, exposing an overlooked vulnerability in diffusion unlearning.
\item We propose \textsc{Tracer}, an efficient framework that accurately identifies multiple erased concepts from large candidate vocabularies without generating or classifying images, making broad searches for undisclosed erasure targets practical.
\item \textsc{Tracer} automatically estimates how many concepts were erased, without labeled calibration examples, knowledge of the unlearning algorithm, or information about which weights were modified.
\item Experiments across text-to-image and text-to-video backbones and diverse unlearning methods show that \textsc{Tracer} accurately identifies erased concepts and estimates their number from large candidate vocabularies in seconds, achieving $\boldsymbol{150}$--$\boldsymbol{137{,}000\times}$ and $\boldsymbol{133}$--$\boldsymbol{20{,}000\times}$ speedups over MIA and brute-force search on image and video models, respectively.
\end{itemize}


\section{Related Work}
\label{related}

\subsection{Concept unlearning in diffusion models.}
\label{related_unlearn}

Fine-tuning methods steer target-concept noise toward anchor distributions~\citep{gandikota2023erasing,kumari2023ablating,zhang2024forget,heng2023selective,fan2023salun,lyu2024one,lu2024mace,zhao2024separable,huang2024receler,bui2025fantastic,thakral2025fine,li2025one,wu2025erasing}.
Closed-form methods edit attention projections without retraining~\citep{gandikota2024unified,orgad2023editing,basu2024localizing,chavhan2024conceptprune,li2025speed,wang2025precise,deng2026forget,deng2026forget1,yaoerasing,biswas2026cure,lee2025localized}.
Inference-time methods suppress undesirable concepts via guidance or activation intervention~\citep{schramowski2023safe,chen2025growth,deja2025saeuron,zhang2025concept}.
%
Recent work extends erasure to flow-matching transformers, including SD3 and FLUX~\citep{gao2025eraseanything,ranjan2026razor,jiang2026z,lee2026localized}.
%
Parallel work targets text-to-video models through editing, low-rank refusal vectors, and training-free safeguards~\citep{liu2024unlearning,yoon2025safree,ye2025t2vunlearning,xu2025videoeraser,facchiano2026video}.
%
Robust methods resist adversarial prompts and subsequent fine-tuning~\citep{zhang2024defensive,kim2024race,gong2024reliable,srivatsan2025stereo,bui2024erasing,li2026aegis,gao2025meta,li2026towards}.

\subsection{Adversarial vulnerabilities of unlearned models.}
\label{re_atk}

Recent analyses show erased concepts are often suppressed, not removed, resurfacing under subsequent benign fine-tuning or alternative sampling strategies~\citep{rusanovsky2025memories,suriyakumar2024unstable,lu2026concepts,yuan2025lurks}.
Adversarial attacks recover erased concepts. White-box attacks exploit model gradients or internal signals~\citep{yang2024mma,chin2023prompting4debugging,zhang2024generate,pham2023circumventing}, whereas black-box attacks search adversarial prompts with query-only model access~\citep{tsai2024ring,lyu2025pla,yook2025zium}.
Non-prompt inference-time attacks recover concepts via concept arithmetic, image-guided inversion, or text-free inversion~\citep{petsiuk2024concept,liu2026image,xiang2026tina}, while weight-space attacks use fine-tuning or pre-erasure weights~\citep{george2025illusion,horwitz2024recovering}.

Beyond unlearning attacks, diffusion models themselves also face classical privacy attacks on their original training data.
Membership inference attacks determine whether a given sample actually belongs to the training set~\citep{matsumoto2023membership,duan2023diffusion,kong2024efficient,fu2023probabilistic,wu2022membership}.
Data extraction and memorization attacks reconstruct or replicate individual training images from the model~\citep{carlini2023extracting,somepalli2023diffusion,somepalli2023understanding}.

\subsection{Positioning of our study.}
Our study aims to identify the concepts targeted for erasure in an unlearned model.
Existing attacks~(Sec~\ref{re_atk}) primarily recover content associated with known erased concepts.
\textsc{Tracer} identifies unknown erasure targets and estimates their number without image generation, fine-tuning, or knowledge of the erasure algorithm.
This setting shifts the focus from attacking a model to regenerate known erased concepts to identifying which concepts were erased.

\section{Method}
\label{method}




\subsection{Unlearn Threat Model}
\label{sec:threat}

\paragraph{What the adversary holds.}
A publisher starts from public base weights $W$, applies a published concept-unlearning method~(Section~\ref{related_unlearn}) to erase an unknown set $S^{\star}$ of concepts, and releases the resulting edited weights $W'$.
The adversary has access only to $W$, $W'$, and a vocabulary $\mathcal{C}$ of ordinary words assumed to contain $S^{\star}$.
In particular, the adversary knows neither $|S^{\star}|$ nor which erasure algorithm was used, has no labelled examples of erased or non-erased models, and therefore cannot construct shadow models for validation.
The adversary may evaluate the full computation graph on text inputs and record the inputs to all changed weight matrices, which are identified directly by comparing $W$ and $W'$.
Text-encoder matrices are probed from the prompt alone, whereas generative-backbone matrices are probed by pairing the prompt with a single fixed latent at several noise levels.
Both probes require only the available model weights and architecture.
The sampling loop is never executed, no image is decoded, and no model output is observed.

\paragraph{What the attack recovers.}
The goal of the attack is to recover the identities of the erased concepts: a subset $\hat{S} \subseteq \mathcal{C}$ that should coincide with $S^{\star}$, together with an inferred count $n$ intended to match $|S^{\star}|$.
Because \textsc{Tracer} automatically searches all available aligned model matrices for weight changes and reads the representations they receive, it requires no knowledge of either the erasure algorithm or its edit locations.
It therefore applies with minimal assumptions to weight-editing concept-unlearning methods across different model components and architectures.

\subsection{Erasure Footprint Assembly}
\label{sec:assembly}

\begin{figure}[t]
\centering
\includegraphics[width=\linewidth]{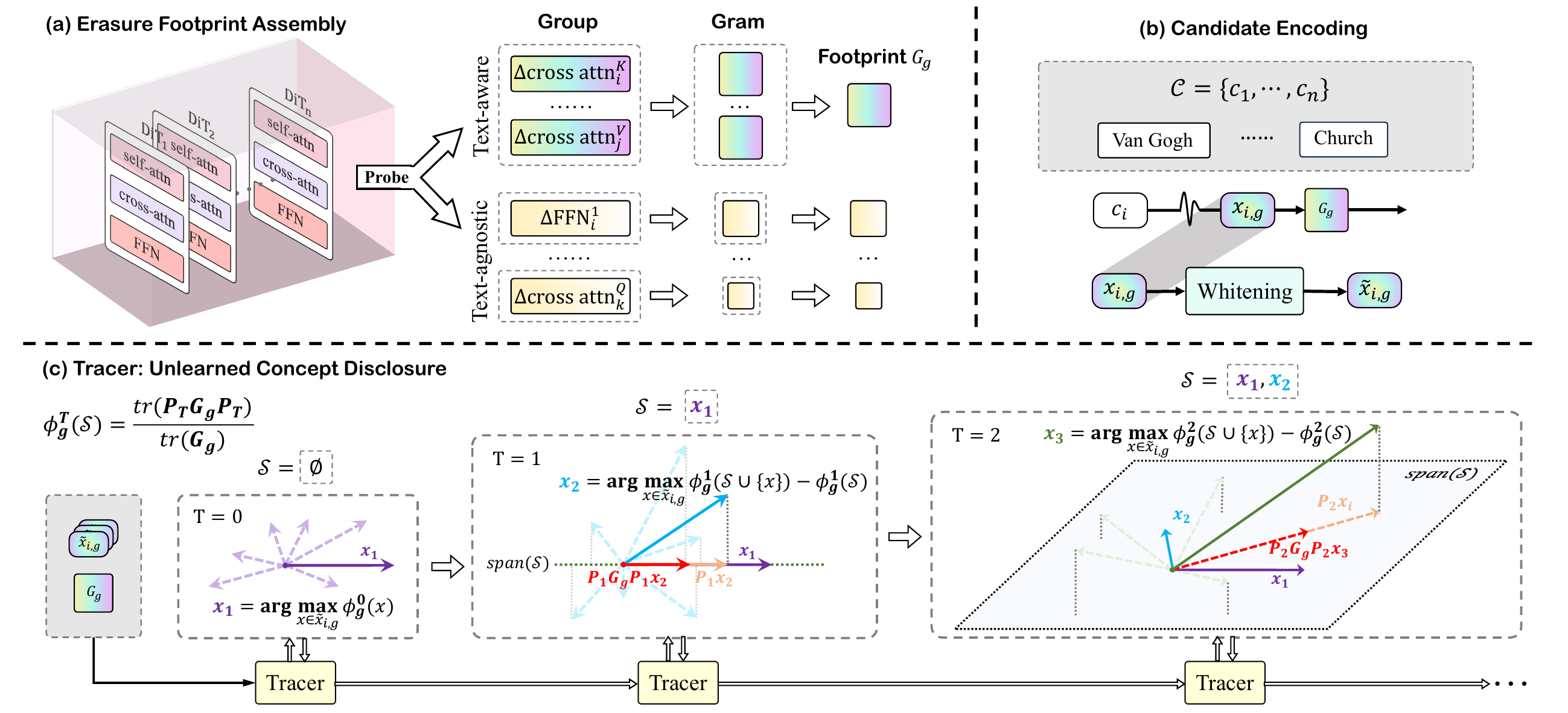}
\caption{\textbf{Overview of \textsc{Tracer}.}
\textbf{(a) Erasure footprint assembly.}
Erasure imprints are grouped by their probed input representations.
Spectrally reshaped and reweighted Gram matrices are fused within compatible groups into footprints $G_g$.
\textbf{(b) Candidate encoding.}
Candidates are encoded in each group's input space, whitened to suppress shared structure, and normalized.
\textbf{(c) Unlearned concept disclosure.}
\textsc{Tracer} iteratively selects concepts explaining the remaining footprint energy. selection combines standardized gains across groups with confidence-based weights.
Label-free self-calibration fixes parameters and group weights.
Confidence drops determine set size without image generation or algorithm knowledge.}
\label{fig:pipeline}
\vspace{-14pt}
\end{figure}

Erasure locations depend on the unlearning method and architecture:
closed-form editors typically modify cross-attention key and value projections,
while fine-tuning may affect other attention and feed-forward layers,
including joint or temporal attention in diffusion-transformer and video backbones.
\textsc{Tracer} identifies changes from weight differences
throughout the computation graph,
without knowledge of edit locations.
For each matrix $\ell$, its \emph{imprint} is
$\Delta W_\ell=W'_\ell-W_\ell$; collectively, non-negligible imprints
form the erasure \emph{\textbf{footprint}}.
We retain $\ell$ in $\mathcal L$ when its relative Frobenius change
$\|\Delta W_\ell\|_F/\|W_\ell\|_F$ exceeds $\tau=10^{-8}$.
This threshold excludes numerical noise.
It is four orders of magnitude below the smallest relative imprint
observed in our experiments.

Combining these imprints introduces two challenges.
\textbf{First}, imprints may receive different representations:
some act on text encodings; others receive internal activations
with different dimensions and semantics.
We identify these representations using probe words rather than layer names.
An imprint is \emph{text-aware} if, for every probe, its input has
greater cosine similarity to the matching text encoding than to
any other probe encoding.
%
This correspondence remains exact for imprints acting on text encodings
but can disappear as textual information mixes with image
features.
Text-aware imprints share one representation group.
Each imprint failing the test is \emph{text-agnostic} and forms a
singleton group with location-specific input.
Only imprints within the same group are aggregated.

\textbf{Second}, imprints within a group may have different output widths
despite sharing an input representation, preventing direct summation.
The Gram matrix $M_\ell=\Delta W_\ell^\top\Delta W_\ell$ removes the
output dimension while preserving imprint strength along the shared
input space, making these matrices summable.
Direct summation may let strongly edited matrices obscure weaker
but informative imprints.
We therefore reshape each Gram matrix before aggregation.
Writing $M_\ell=Q\Lambda Q^\top$, the eigenvectors in $Q$ specify its
principal directions, and each eigenvalue $\lambda$ measures the
corresponding strength.
Reshaping preserves these directions, raises eigenvalues normalized
by $\lambda_{\max}$ to a power $\rho$, and discards negligible directions.
We additionally weight imprints by their inverse edit magnitude,
with strength controlled by $\beta$:
\begin{equation}
G_g
=
\sum_{\ell\in g}
\|\Delta W_\ell\|_F^{-2\beta}
\bigl[\!\bigl[M_\ell\bigr]\!\bigr]_\rho,
\qquad
\bigl[\!\bigl[M_\ell\bigr]\!\bigr]_\rho
=
Q\,\mathrm{diag}\!\Bigl(
(\lambda/\lambda_{\max})^\rho
\mathbf{1}\!\left[\lambda/\lambda_{\max}\geq\varepsilon_0\right]
\Bigr)Q^\top.
\label{eq:fusion}
\end{equation}
Here, $\mathbf{1}[\cdot]$ denotes the indicator function,
and $\varepsilon_0$ is the relative spectral cutoff.
The self-calibrated exponents $\rho$ and $\beta$ control the relative
strengths of directions within each imprint and contributions across
imprints, respectively.
This produces one \emph{aggregated footprint} $G_g$ per representation group.

\subsection{Geometry-Aware Candidate Encoding}
\label{sec:encoding}

Each vocabulary entry in $\mathcal C$ is a \emph{candidate}.
We encode each candidate in the input space of its group's footprint.
We insert the candidate into a set $T$ of neutral prompt templates,
pass the prompts through the network, and average the representations
received by group $g$ across templates to obtain $x_{c,g}$.
One template turns \emph{church} into ``a photo of a church.''
For a text-aware group, we use the text encoding at the token position
specified by offset $o$.
For a text-agnostic group, we use the internal activation at that location,
averaged over applicable token positions.
Groups in the generative backbone additionally average over fixed
noise levels.
Template averaging reduces sensitivity to phrasing and emphasizes the
candidate.
Each candidate is represented separately in each group's input space.

Candidate representations are unevenly distributed within each group's input
space: a few directions carry much of the vocabulary energy, giving many
candidates large shared components.
Candidates aligned with these directions may score highly against different
footprints regardless of which concepts were erased.
We suppress this shared structure to emphasize candidate-specific information.
We construct a regularized second-moment matrix over the full vocabulary
and normalize the transformed representations:
\begin{equation}
\Sigma_g
=
\mathbb{E}_{c\in\mathcal C}
\bigl[x_{c,g}x_{c,g}^{\top}\bigr]
+\varepsilon_w I,
\qquad
\tilde{x}_{c,g}
=
\frac{\Sigma_g^{-\gamma/2}x_{c,g}}
{\bigl\|\Sigma_g^{-\gamma/2}x_{c,g}\bigr\|},
\label{eq:whiten}
\end{equation}
where the expectation is the uncentered second moment and $I$ is the identity.
We set
$\varepsilon_w=10^{-3}\mathbb{E}_{c\in\mathcal C}\|x_{c,g}\|^2/d_g$,
where $d_g$ is the group's input dimension.
Regularization ensures positive definiteness and limits amplification
of weak directions.
We compute $\Sigma_g^{-\gamma/2}$ through eigendecomposition.
Directions carrying more vocabulary energy are attenuated more strongly.
$\gamma$ controls suppression from no transformation
to full whitening.
$o$ and $\gamma$ are selected by self-calibration in
Section~\ref{sec:disclosure} before iterative candidate selection.
Unit normalization prevents larger norms from inflating candidate scores
in footprint comparisons using $\tilde{x}_{c,g}$.

\subsection{Unlearned Concept Disclosure}
\label{sec:disclosure}

\paragraph{Subspace Coverage Scoring.}
The footprint identifies a subspace whose eigenvectors may rotate within
eigenspaces with equal or nearly equal eigenvalues.
We therefore match candidate spans to the footprint rather than associate
individual spectral directions with words.
For $S\subseteq\mathcal C$, coverage measures the fraction of footprint
energy captured by its candidate span:
\begin{equation}
\phi_g(S)
=
\frac{\mathrm{tr}\bigl(P_{S,g}G_gP_{S,g}\bigr)}
{\mathrm{tr}(G_g)},
\label{eq:coverage}
\end{equation}
where $P_{S,g}$ projects orthogonally onto
$\operatorname{span}\{\tilde{x}_{c,g}:c\in S\}$ and
$\mathrm{tr}(\cdot)$ denotes the trace.
The singleton score $s_g(c)=\phi_g(\{c\})$ defines the group's initial field.
Additional coverage measures contributions beyond $S$,
with zero gain for candidates within the span.
To measure field separation, we define
\begin{equation}
\mathrm{conf}_\kappa(s)
=
\frac{\max_c s(c)-\mathrm{med}(s)}
{q_{1-\kappa}(s)-\mathrm{med}(s)+\varepsilon_c}.
\label{eq:conf}
\end{equation}
Here, $s$ is the field over candidates,
$\mathrm{med}(s)$ its median, and $q_{1-\kappa}(s)$ its
$(1-\kappa)$-quantile.
$\kappa\in(0,1/2)$ specifies the upper-tail comparison range,
while $\varepsilon_c>0$ stabilizes the ratio.
A constant field has zero confidence.
The numerator measures the leader's separation from the median;
the denominator normalizes it by the upper-tail spread.
This reduces sensitivity to score scale while retaining dependence
on the candidate vocabulary.
Groups observe different representations, so we standardize their gains
and weight them by the separation of leading candidates in their initial fields:
\begin{equation}
z_g(c,S)
=
\frac{\phi_g\bigl(S\cup\{c\}\bigr)-\phi_g(S)-\mu_g(S)}
{\sigma_g(S)+\varepsilon_z},
\qquad
w_g
=
\frac{\max\{\mathrm{conf}_\kappa(s_g),\varepsilon_g\}}
{\sum_{g'}\max\{\mathrm{conf}_\kappa(s_{g'}),\varepsilon_g\}},
\label{eq:merge}
\end{equation}
where $\mu_g(S)$ and $\sigma_g(S)$ are the gain mean and standard
deviation over $\mathcal C$.
The constants $\varepsilon_z>0$ and $\varepsilon_g>0$ stabilize
standardization and ensure positive group weights, respectively.
For any parameter setting, the initial combined field is
$s^{(1)}(c)=\sum_g w_g z_g(c,\varnothing)$.

\paragraph{Self-Calibration.}
Candidate fields depend on spectral shaping $\rho$, whitening $\gamma$,
imprint-amplitude weighting $\beta$, and the token offset $o$.
Without a labelled validation set, the adversary evaluates each setting
through its initial combined field, before selecting a candidate.
Writing $\theta=(\beta,\gamma,\rho,o)$, we evaluate a predefined
finite set $\Theta$ and select
$\theta^\star=\operatorname*{arg\,max}_{\theta\in\Theta}
\mathrm{conf}_\kappa(s_\theta^{(1)})$.
$s_\theta^{(1)}$ is constructed with the footprint operators,
candidate representations, and group weights induced by $\theta$.
This selection requires no knowledge of the erased-set size.
The setting and its group weights remain fixed throughout
the search.
Candidate gains and their standardization are updated as the
span expands.

\paragraph{Progressive Concept Pursuit.}
With the calibrated setting, we greedily construct a candidate sequence
by maximizing additional coverage across groups.
Overlapping vocabulary expressions may remain prominent after a related
candidate is selected because their representations need not lie in its span.
We suppress these lexical overlaps after each selection.
Let $\mathcal F(c)$ contain $c$ and all vocabulary entries whose token
sets are subsets or supersets of its token set.
These entries are excluded from subsequent selection using only the
vocabulary, without erased-concept labels or an assumption of semantic equivalence.
Starting from $S_0=D_0=\varnothing$, let $S_j$ and $D_j$ denote the
selected and excluded entries after step $j$.
Each step combines the group gains into the global score
$s^{(j)}(c)=\sum_g w_g z_g(c,S_{j-1})$.
We select the candidate $c_j$ with the highest score among entries
not in $D_{j-1}$.
We add $c_j$ to the selected set and exclude all entries in
$\mathcal F(c_j)$ from subsequent selection.
Other suppressed entries are not added to the selected span.
The search uses a fixed horizon $J$ without prior knowledge of the
erased-set size.
We record confidence over eligible candidates initially and after
each selection to determine how many selected candidates are retained
in the output.

\paragraph{Erased-Set Size Detection.}
We determine the candidate sequence's cutoff by tracking how clearly
the remaining field separates its leader from the background.
A pronounced decline suggests exhaustion of the dominant erasure signal.
After selecting $j$ candidates, we record
$a_j=\mathrm{conf}_\kappa(s^{(j+1)})$ for $j=0,\ldots,J$.
Confidence is computed only over candidates that have neither been
selected nor excluded by lexical suppression.
Thus, $a_0$ measures initial separation, while $a_J$ measures separation
after the final selection.
We measure the confidence decline following the $i$-th selection as
\begin{equation}
r_i
=
\log\max\{a_{i-1},\varepsilon_a\}
-
\log\max\{a_i,\varepsilon_a\},
\qquad i=1,\ldots,J,
\label{eq:confdrop}
\end{equation}
where $\varepsilon_a>0$ keeps logarithms defined.
A positive $r_i$ indicates a decline, with its magnitude measuring
the log confidence ratio.
The largest decline provides the default size estimate.
Some profiles rise as selecting dominant candidates makes remaining
targets more distinguishable, then decline in stages because related
vocabulary entries remain prominent.
The largest decline may therefore occur after the erased set
has been selected.
For these profiles, we identify the first sufficiently large decline.
The initial-rise condition holds when confidence increases in at least
$B-1$ of the first $B$ transitions, where $3\leq B\leq J$
is a fixed window length.
Let $r_{\max}$ denote the largest log confidence drop
and $\alpha\in(0,1)$ a relative threshold.
If the initial-rise condition holds and $r_{\max}>0$, we estimate
$n=\min\{i:r_i\geq\alpha r_{\max}\}$ over $i=1,\ldots,J$,
selecting the first drop reaching a fraction $\alpha$ of the largest drop.
Otherwise, $n$ is the index of the largest drop, with ties resolved
by choosing the earliest selection.
Both rules use only the confidence sequence.
The estimate $n$ determines the set $\hat S=S_n$,
comprising the first $n$ candidates.
The horizon $J$ bounds the search without prescribing the output size.
Implementation settings are detailed in Appendix~\ref{app:hyperparameters} and Appendix~\ref{app:multiword}.

\begin{table}[t]
\captionsetup{font=footnotesize,labelfont=bf}
\centering
\caption{Identification accuracy (\%) of erased object concepts
across three text-to-image backbones on Imagenette-20.
Shaded cells report evaluated results.
``--'' indicates an unevaluated setting.
}
\label{tab:k-t2i}

\scriptsize
\renewcommand{\arraystretch}{0.7}
\setlength{\tabcolsep}{1pt}

\def\venue#1{{\sffamily #1}}
\def\NA{--}
\def\result#1{\cellcolor{gray!20}#1}

\begin{tabularx}{\linewidth}{@{}ll*{6}{>{\centering\arraybackslash}X}@{}}
\toprule
\multirow{2}{*}{\textbf{Method}}
& \multirow{2}{*}{\textbf{Venue}}
& \multicolumn{5}{c}{\textbf{Erased-set size $K$}}
& \multirow{2}{*}{\textbf{Avg.} ($\uparrow$)} \\
\cmidrule(lr){3-7}
& & 1 & 3 & 5 & 10 & 20 & \\
\midrule
\multicolumn{8}{c}{\textit{Stable Diffusion v1.5}} \\
\midrule
\textbf{ESD}
& \venue{ICCV~\citeyear{gandikota2023erasing}}
& \result{100} & \result{100} & \result{94}
& \NA & \NA & 98.0 \\

\textbf{UCE}
& \venue{WACV~\citeyear{gandikota2024unified}}
& \result{100} & \result{100} & \result{100}
& \result{100} & \result{100} & 100.0 \\

\textbf{MACE}
& \venue{CVPR~\citeyear{lu2024mace}}
& \result{100} & \result{100} & \result{100}
& \result{100} & \NA & 100.0 \\

\textbf{FMN}
& \venue{CVPR~\citeyear{zhang2024forget}}
& \result{100} & \result{100} & \result{100}
& \NA & \NA & 100.0 \\

\textbf{ConceptPrune}
& \venue{ICLR~\citeyear{chavhan2024conceptprune}}
& \result{100} & \result{100} & \result{100}
& \result{94} & \NA & 98.5 \\

\textbf{Sculpting}
& \venue{ICCV~\citeyear{li2025sculpting}}
& \result{100} & \result{100} & \result{100}
& \result{100} & \NA & 100.0 \\

\textbf{SPEED}
& \venue{ICLR~\citeyear{li2025speed}}
& \result{100} & \result{100} & \NA
& \NA & \NA & 100.0 \\

\textbf{ScaPre}
& \venue{ICLR~\citeyear{deng2026forget}}
& \result{100} & \result{100} & \result{100}
& \result{100} & \result{100} & 100.0 \\

\textbf{FIA}
& \venue{ICML~\citeyear{deng2026forget1}}
& \result{100} & \result{100} & \result{100}
& \result{100} & \NA & 100.0 \\
\midrule
\textbf{AdvUnlearn}
& \venue{NeurIPS~\citeyear{zhang2024defensive}}
& \result{100} & \NA & \NA
& \NA & \NA & 100.0 \\

\textbf{STEREO}
& \venue{CVPR~\citeyear{srivatsan2025stereo}}
& \result{100} & \result{100} & \NA
& \NA & \NA & 100.0 \\

\textbf{Meta-Unlearning}
& \venue{ICCV~\citeyear{gao2025meta}}
& \result{100} & \NA & \NA
& \NA & \NA & 100.0 \\
\midrule
\multicolumn{8}{c}{\textit{Stable Diffusion~3}} \\
\midrule
\textbf{UCE}
& \venue{WACV~\citeyear{gandikota2024unified}}
& \result{100} & \result{100} & \result{94}
& \NA & \NA & 98.0 \\

\textbf{EraseFlow}
& \venue{NeurIPS~\citeyear{kusumba2026eraseflow}}
& \result{100} & \result{100} & \result{100}
& \NA & \NA & 100.0 \\
\midrule
\multicolumn{8}{c}{\textit{FLUX.1}} \\
\midrule
\textbf{EraseAnything}
& \venue{ICML~\citeyear{gao2025eraseanything}}
& \result{100} & \result{100} & \NA
& \NA & \NA & 100.0 \\

\textbf{ScaPre}
& \venue{ICLR~\citeyear{deng2026forget}}
& \result{100} & \result{100} & \result{100}
& \result{100} & \NA & 100.0 \\
\bottomrule
\end{tabularx}
\vspace{-1em}
\end{table}

\section{Experiment}
\label{exp}
\label{sec:exp}

\subsection{Experimental Settings}
\label{sec:exp-setup}

We evaluate \textsc{Tracer} on three text-to-image backbones: Stable Diffusion
v1.5~\citep{rombach2022high}, Stable Diffusion~3~\citep{esser2024scaling}, and
FLUX.1~\citep{flux2024}.
The text-to-video backbones are CogVideoX-5B~\citep{yang2025cogvideox} and
Wan2.1-T2V-1.3B~\citep{wan2025}.
We use official implementations and released checkpoints without modification.
Object identification is evaluated on the ten Imagenette classes~\citep{imagenette}.
Explicit-content evaluation uses I2P prompts~\citep{schramowski2023safe}.
Artistic-style identification covers the artists evaluated by
ESD~\citep{gandikota2023erasing} and UCE~\citep{gandikota2024unified}.
Generation quality is evaluated on MS COCO~\citep{lin2014microsoft}.
We construct \textbf{Imagenette-20} by augmenting Imagenette's ten
classes~\citep{imagenette} with 10 non-overlapping common object categories from ImageNet-1k~\citep{russakovsky2015imagenet} to evaluate identification across varying erased-set sizes.
All experiments run on NVIDIA RTX A6000 GPUs. 
%

\subsection{Disclosing Erased Concept in Unlearned Models}

\label{sec:exp-main}

Reproducing unlearning methods with their official implementations
reveals limits on stable erasure capacity.
Beyond these limits, models exhibit degraded forgetting or
damaged generative capabilities.
Shaded cells in Table~\ref{tab:k-t2i} indicate evaluated settings
within these limits; excluded settings are marked ``--''.
We evaluate varying erased-set sizes, where $K$ denotes
the number of concepts jointly erased from each checkpoint.
For each size, we sample concept sets from Imagenette-50,
shared across methods and backbones.
Each set is jointly erased to produce a checkpoint.
We define \emph{recovery accuracy} as the percentage of checkpoints
whose recovered set exactly matches the ground-truth erased set.
Correct recovery identifies all erased concepts without extras,
jointly testing concept identities and inferred set size.
Partial recovery counts as failure.
The Avg.\ column averages recovery accuracy over evaluated
set sizes within each method's stable erasure range, excluding ``--''.
Higher values indicate better performance ($\uparrow$).
%
Table~\ref{tab:k-t2v} extends the same protocol to text-to-video models.
Evaluation is restricted to individual concepts and small groups,
where the tested methods maintain stable erasure.
\textsc{Tracer} achieves perfect recovery for both methods
on CogVideoX-5B and for Refusal on Wan2.1-T2V-1.3B.
For VideoEraser on Wan2.1-T2V-1.3B, recovery remains perfect
for individual concepts, with a modest decline under joint erasure.
These results demonstrate that erasure footprints remain informative
across video architectures with different temporal modeling,
enabling accurate identification of both individual targets
and jointly erased concept sets.

\begin{figure}[t]
\centering

\begin{minipage}[t]{0.46\textwidth}
\vspace{0pt}
\captionsetup{type=table,font=footnotesize,labelfont=bf,
             position=top,skip=4pt}
\caption{Exact estimation accuracy (\%) of the erased-set
size $K$ across evaluated image and video unlearning methods.}
\label{tab:rank-detection}
\centering

\scriptsize
\renewcommand{\arraystretch}{0.7}
\setlength{\tabcolsep}{1pt}

\def\result#1{\cellcolor{gray!20}#1}

\begin{tabularx}{\linewidth}{@{}l*{5}{>{\centering\arraybackslash}X}@{}}
\toprule
\multirow{2}{*}{\textbf{Method}}
& \multicolumn{5}{c}{\textbf{Ground-truth size $K$}} \\
\cmidrule(lr){2-6}
& 1 & 3 & 5 & 10 & 20 \\
\midrule
\multicolumn{6}{c}{\textit{text-to-image} (\textit{Stable Diffusion v1.5})} \\
\midrule
\textbf{ESD}
& \result{100} & \result{100} & \result{100}
& -- & -- \\
\textbf{UCE}
& \result{100} & \result{100} & \result{100}
& \result{100} & \result{100} \\
\textbf{MACE}
& \result{100} & \result{100} & \result{100}
& \result{100} & -- \\
\textbf{ConceptPrune}
& \result{100} & \result{100} & \result{98}
& -- & -- \\
\textbf{SPEED}
& \result{100} & \result{100} & \result{92}
& -- & -- \\
\textbf{ScaPre}
& \result{100} & \result{100} & \result{100}
& \result{100} & \result{100} \\
\midrule
\multicolumn{6}{c}{\textit{text-to-image} (\textit{FLUX.1})} \\
\midrule
\textbf{EraseAnything}
& \result{100} & \result{100} & -- & -- & -- \\
\textbf{EraseFlow}
& \result{100} & \result{100} & \result{100}
& -- & -- \\
\midrule
\multicolumn{6}{c}{\textit{text-to-video} (\textit{CogVideoX-5B})} \\
\midrule
\textbf{Refusal}
& \result{100} & \result{100} & \result{100}
& -- & -- \\
\textbf{VideoEraser}
& \result{100} & \result{100} & -- & -- & -- \\
\bottomrule
\end{tabularx}
\end{minipage}
\hfill
\begin{minipage}[t]{0.50\textwidth}
\vspace{0pt}
\centering
\captionsetup{type=figure,font=footnotesize,labelfont=bf,
             position=bottom,skip=0pt}

\includegraphics[width=\linewidth]{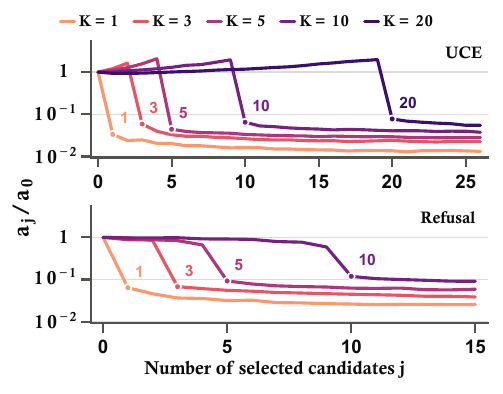}
\caption{Normalized confidence $a_j/a_0$ drops at the true
erased-set size $K$ for UCE and Refusal.}
\label{fig:rank-eigengap}
\end{minipage}

\end{figure}

To assess size estimation separately from concept identification,
Table~\ref{tab:rank-detection} reports the percentage of checkpoints
whose inferred count matches the true erased-set size.
\textsc{Tracer} achieves perfect size estimation in most settings
across image and video models, with occasional errors
for ConceptPrune and SPEED under joint erasure.
Figure~\ref{fig:rank-eigengap} illustrates the confidence dynamics
for \textsc{UCE} and \textsc{Refusal}.
Normalized confidence remains high during early selections,
then drops sharply once the selected sequence reaches
the true erased-set size.
This decline reflects reduced separation between the leading
remaining candidate and the background, suggesting exhaustion
of the dominant erasure signal.
The confidence-drop rule uses this transition to estimate
the target count without prior size information.
Artistic-style identification results are reported
in Appendix~\ref{app:results}.

\begin{table}[t]
\centering

\begin{minipage}[t]{0.59\linewidth}
\vspace{0pt}
\captionsetup{type=table,font=footnotesize,labelfont=bf,
             position=top,skip=4pt}
\caption{Recall@$K$ (\%) with the true $K$ provided,
evaluated over 50 checkpoints per setting.
}
\label{tab:auditing-comparison}
\centering

\scriptsize
\renewcommand{\arraystretch}{0.7}
\setlength{\tabcolsep}{1pt}

\begin{tabularx}{\linewidth}{@{}ll*{4}{>{\centering\arraybackslash}X}@{}}
\toprule
\multirow{2}{*}{\textbf{Baseline}}
& \multirow{2}{*}{\textbf{Method}}
& \multicolumn{3}{c}{\textbf{Recall@$K$} ($\uparrow$)}
& \multirow{2}{*}{\shortstack{\textbf{Recall@10}\\($\uparrow$)}} \\
\cmidrule(lr){3-5}
& & $K=1$ & $K=3$ & $K=5$ & \\
\midrule
\multirow{3}{*}{\textbf{ESD}}
& MIA
& 42.0 & 6.7 & 4.8 & 27.1 \\
& Brute-force
& 82.0 & 33.3 & 16.2 & 53.3 \\
& \textsc{Tracer}
& \textbf{100.0} & \textbf{100.0}
& \textbf{100.0} & \textbf{100.0} \\
\midrule
\multirow{3}{*}{\textbf{UCE}}
& MIA
& 60.0 & 24.0 & 14.0 & 39.6 \\
& Brute-force
& 92.0 & 68.7 & 36.6 & 72.2 \\
& \textsc{Tracer}
& \textbf{100.0} & \textbf{100.0}
& \textbf{100.0} & \textbf{100.0} \\
\midrule
\multirow{3}{*}{\textbf{SPEED}}
& MIA
& 78.0 & 63.3 & 42.4 & 68.9 \\
& Brute-force
& 76.0 & 60.0 & 38.8 & 66.5 \\
& \textsc{Tracer}
& \textbf{100.0} & \textbf{100.0}
& \textbf{100.0} & \textbf{100.0} \\
\midrule
\multirow{3}{*}{\textbf{ScaPre}}
& MIA
& 80.0 & 72.0 & 48.2 & 74.7 \\
& Brute-force
& 90.0 & 78.7 & 52.0 & 80.4 \\
& \textsc{Tracer}
& \textbf{100.0} & \textbf{100.0}
& \textbf{100.0} & \textbf{100.0} \\
\bottomrule
\end{tabularx}
\end{minipage}
\hfill
\begin{minipage}[t]{0.39\linewidth}
\vspace{0pt}
\begingroup
\setcitestyle{numbers,square}
\captionsetup{type=table,font=footnotesize,labelfont=bf,
             position=top,skip=4pt}
\centering
\caption{Identification accuracy (\%) of erased object concepts
across two text-to-video backbones with $K\in\{1,3\}$,
following the protocol of Table~\ref{tab:k-t2i}.}
\label{tab:k-t2v}

\scriptsize
\renewcommand{\arraystretch}{1.1}
\setlength{\tabcolsep}{1pt}

\def\venue#1{{\sffamily #1}}

\begin{tabularx}{\linewidth}{@{}ll*{3}{>{\centering\arraybackslash}X}@{}}
\toprule
\multirow{2}{*}{\textbf{Method}}
& \multirow{2}{*}{\textbf{Venue}}
& \multicolumn{2}{c}{\textbf{Size $K$}}
& \multirow{2}{*}{\shortstack{\textbf{Avg.}\\($\uparrow$)}} \\
\cmidrule(lr){3-4}
& & 1 & 3 & \\
\midrule
\multicolumn{5}{c}{\textit{CogVideoX-5B}} \\
\midrule
\textbf{VideoEraser}
& \venue{EMNLP~\citeyear{xu2025videoeraser}}
& 100 & 100 & 100.0 \\
\textbf{Refusal}
& \venue{ICLR~\citeyear{facchiano2026video}}
& 100 & 100 & 100.0 \\
\midrule
\multicolumn{5}{c}{\textit{Wan2.1-T2V-1.3B}} \\
\midrule
\textbf{VideoEraser}
& \venue{EMNLP~\citeyear{xu2025videoeraser}}
& 100 & 90 & 95.0 \\
\textbf{Refusal}
& \venue{ICLR~\citeyear{facchiano2026video}}
& 100 & 100 & 100.0 \\
\bottomrule
\end{tabularx}
\endgroup
\end{minipage}

\end{table}

\subsection{Comparison With Traditional Adversarial Methods}
\label{sec:exp-baseline}

We compare \textsc{Tracer} with two approaches.
Adapted MIA ranks concepts by aggregated changes in sample-level
membership scores.
Brute-force probing ranks candidates by comparing images generated
before and after unlearning.
Implementation details are provided in Appendix~\ref{app:base_impl}.
We evaluate ESD, UCE, SPEED, and ScaPre on Stable Diffusion v1.5,
with 50 checkpoints per erased-set size $K\in\{1,3,5\}$.
All approaches share checkpoints and 1,000-name vocabulary.
To accommodate conventional methods, all receive the true $K$;
\textsc{Tracer} uses its iterative selection order without
size estimation.
Unlike exact-set accuracy, this comparison credits partial recovery.
Let $T_m^{(L)}$ contain the first $L$ candidates
for checkpoint $m$.
We compute
$\mathrm{Recall@}L=\frac{100}{50}\sum_{m=1}^{50}
|T_m^{(L)}\cap S_m^\star|/K$,
where $S_m^\star$ is the erased set.
Table~\ref{tab:auditing-comparison} reports Recall@$K$
for each $K$; recovering three of five targets scores 60\%.
Recall@10 uses $L=10$ with denominator $K$.
Its reported value equally averages the three $K$ settings.
\textsc{Tracer} maintains high recall across methods and set sizes, while baseline recall generally declines as $K$ grows.
Allowing ten candidates improves baseline coverage but
leaves targets unidentified, supporting the effectiveness of
weight footprints beyond behavioral changes alone.

\begin{figure}[t]
\centering
\includegraphics[width=\linewidth]{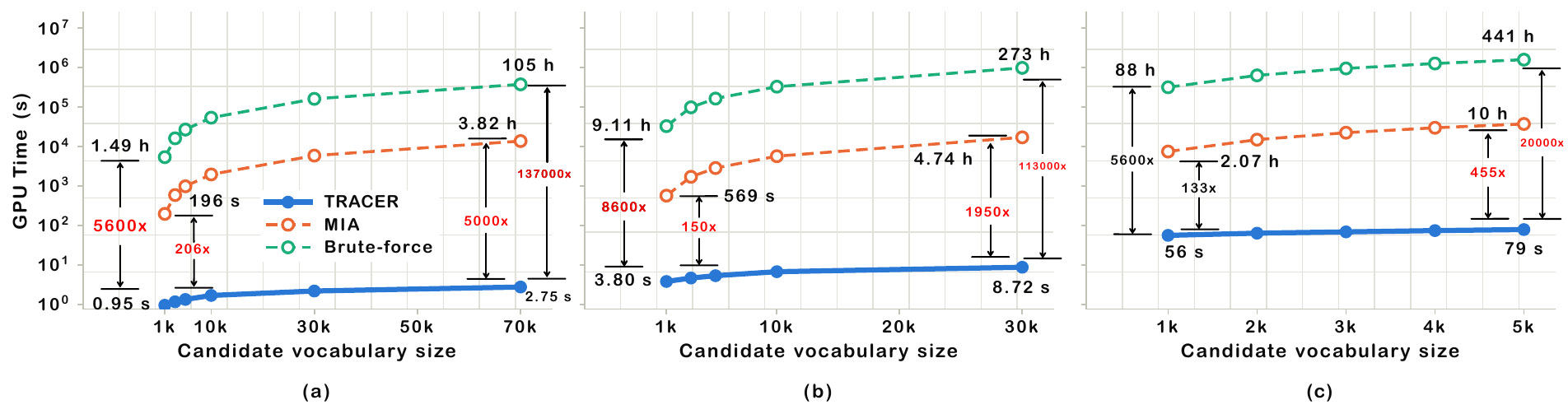}
\caption{Runtime across three methods, showing
the scalability of \textsc{Tracer} to large candidate sets.}
\label{fig:auditing-runtime}
\end{figure}

Figure~\ref{fig:auditing-runtime} compares runtime across candidate
vocabulary sizes on (a) Stable Diffusion v1.5, (b) FLUX.1,
and (c) CogVideoX-5B.
\textsc{Tracer} exhibits modest runtime growth as the vocabulary
expands across all backbones.
Across the evaluated vocabulary sizes, \textsc{Tracer} achieves
$150$--$137{,}000\times$ and $133$--$20{,}000\times$ speedups
over MIA and brute-force probing on image and video models, respectively.
Weight analysis with limited forward probes avoids repeated
image or video generation and sample-level membership evaluation,
enabling efficient concept identification over large candidate vocabularies.

\subsection{Ablation Studies}
\label{sec:exp-ablation}

We evaluate components on 50 checkpoints per unlearning
method at $K=5$.
Table~\ref{tab:ablation-components} provides $K$ to isolate
concept identification from size estimation.
The single imprint variant retains the imprint with the highest
initial confidence in each group.
Removing background suppression sets $\gamma=0$ while preserving
unit normalization.
Removing progressive selection uses the initial ranking without
updating coverage gains, retaining lexical suppression.
\textsc{Tracer} achieves 100\% Recall@5 across four methods.
Using a single imprint reduces recall by 5.2 to 16.0 percentage points,
supporting evidence integration across weight changes.
Removing background suppression causes declines of 7.6 to 23.2 points.
Replacing progressive selection with static ranking produces the
largest reduction across methods, reaching 43.6 points on ESD.
These results support updating candidate scores
to capture information beyond the selected span.
Table~\ref{tab:ablation-calibration} evaluates exact recovery of
the erased set without providing $K$.
Replacing self-calibration with fixed parameters
reduces accuracy to between 78\% and 92\%.
Always selecting the largest confidence drop, without accounting
for an initial rise, yields accuracy between 64\% and 80\%.
The full method achieves 100\% across four methods,
supporting both calibration for each checkpoint and size estimation
adapted to the confidence profile.






\begin{table}[t]
\centering
\captionsetup{font=footnotesize,labelfont=bf,skip=3pt}

\scriptsize
\setlength{\tabcolsep}{2pt}
\renewcommand{\arraystretch}{1.1}
\setlength{\heavyrulewidth}{0.8pt}
\setlength{\lightrulewidth}{0.4pt}
\setlength{\aboverulesep}{1.5pt}
\setlength{\belowrulesep}{1.5pt}

\begin{minipage}[t]{0.49\linewidth}
\vspace{0pt}
\caption{Component ablations in Recall@5 (\%) with the true number of erasure targets $K=5$ provided.}
\label{tab:ablation-components}
\centering

\begin{tabular*}{\linewidth}{@{\extracolsep{\fill}}lcccc@{}}
\toprule
\textbf{Variant} & \textbf{ESD} & \textbf{UCE}
& \textbf{SPEED} & \textbf{ScaPre} \\
\midrule
Full \textsc{Tracer}
& \textbf{100.0} & \textbf{100.0}
& \textbf{100.0} & \textbf{100.0} \\
Single imprint per group
& 84.0 & 94.8 & 88.4 & 91.6 \\
w/o background suppression
& 76.8 & 92.4 & 90.8 & 88.0 \\
w/o progressive selection
& 56.4 & 78.4 & 82.8 & 81.2 \\
\bottomrule
\end{tabular*}
\end{minipage}
\hfill
\begin{minipage}[t]{0.49\linewidth}
\vspace{0pt}
\caption{Calibration and size-detection ablations in
exact-set recovery accuracy (\%).}
\label{tab:ablation-calibration}
\centering

\renewcommand{\arraystretch}{1.38}
\begin{tabular*}{\linewidth}{@{\extracolsep{\fill}}lcccc@{}}
\toprule
\textbf{Variant} & \textbf{ESD} & \textbf{UCE}
& \textbf{SPEED} & \textbf{ScaPre} \\
\midrule
Full \textsc{Tracer}
& \textbf{100.0} & \textbf{100.0}
& \textbf{100.0} & \textbf{100.0} \\
w/o self-calibration
& 78 & 84 & 92 & 86 \\
Largest-drop only
& 72 & 64 & 78 & 80 \\
\bottomrule
\end{tabular*}
\end{minipage}

\end{table}

\section{Conclusion}

We presented \textsc{Tracer}, which accurately identifies erasure targets and their number from publicly released model weights in seconds without image generation.
Experiments demonstrate its effectiveness across image and video backbones and diverse unlearning methods.
It enables label-free search over large candidate vocabularies, achieving substantial speedups over membership inference attacks and brute-force generation-based probing methods.
Our work establishes erasure-target identification as a distinct task and provides a practical and computationally efficient framework for assessing what model weights reveal about erasure targets despite suppressed generation.

\clearpage
\newpage
\section*{AI Use Statement}

In this work, generative AI tools were used solely to correct grammatical errors and improve the clarity of the English writing, without changing the scientific content. They were not used to generate research ideas, develop the methodology, conduct experiments, write or analyze code, interpret results, formulate claims, or create figures and other research artifacts. All scientific content and contributions were developed independently by the authors.

\section*{Ethics Statement}

This work does not involve human subjects, personally identifiable information, private data, or sensitive user information. All models and datasets used in this study are publicly available and are employed solely for academic research. We are not aware of any ethical concerns, conflicts of interest, or issues related to discrimination, privacy, security, or legal compliance arising from this work. The authors follow established responsible research practices.

\section*{Reproducibility Statement}

We have made every effort to ensure the reproducibility of this work. The complete methodological pipeline and all components of the proposed approach are described in detail in the main paper, with additional implementation details provided in the appendix. These descriptions are intended to enable independent reproduction of our method and experimental results. Upon acceptance, we will publicly release the complete source code and accompanying configurations.


\bibliography{main}
\bibliographystyle{preprint}

\newpage
\appendix

\section{Implementation Details and Evaluation Protocols}
\label{app:impl}

\subsection{Hyperparameter Settings and Implementation Details}
\label{app:hyperparameters}

\paragraph{Self-calibration grid.}
We select $\theta=(\beta,\gamma,\rho,o)$ by maximizing the confidence
of the initial combined field, as described in
Section~\ref{sec:disclosure}.
Table~\ref{tab:hyperparameter-grid} lists the candidate values.
The predefined finite set $\Theta$ is their full Cartesian product,
yielding $2\times2\times3\times2=24$ settings for the main configuration.
The selected setting and its group weights remain fixed throughout
the search.
Candidate gains and their standardization are updated as the
selected span expands.

\begin{table}[ht]
\centering
\small
\setlength{\tabcolsep}{5pt}
\caption{Self-calibration search grid for the main configuration.}
\label{tab:hyperparameter-grid}
\begin{tabularx}{\linewidth}{@{}clX@{}}
\toprule
Parameter & Candidate values & Role \\
\midrule
$\beta$ & $\{0,1\}$
& Controls imprint-amplitude weighting through
$\|\Delta W_\ell\|_F^{-2\beta}$.
Setting $\beta=0$ removes amplitude weighting. \\
$\gamma$ & $\{0,1\}$
& Controls background suppression, from no transformation
to full regularized whitening.
Unit normalization is retained in both cases. \\
$\rho$ & $\{0,0.5,1\}$
& Controls spectral shaping within each imprint. \\
$o$ & $\{2,1\}$
& Specifies the token position relative to the non-padding
sequence length for token-based text encodings. \\
\bottomrule
\end{tabularx}
\end{table}

For $\rho=0$, spectral reshaping produces the orthogonal projector
onto the eigenspace retained by the relative spectral cutoff
$\varepsilon_0$ in Eq.~\eqref{eq:fusion}.
For $\rho=1$, it retains the normalized eigenvalues
$\lambda/\lambda_{\max}$ above this cutoff.
The intermediate setting $\rho=0.5$ reduces the disparity
between the strengths of retained spectral directions.

\paragraph{Token Position Selection.}
For a text-aware group using token-based text encodings,
the offset $o$ specifies the token position used to represent
each prompt, as described in Section~\ref{sec:encoding}.
Let $L(p)$ denote the number of non-padding tokens in prompt $p$,
including special tokens.
The zero-based token index is
\begin{equation}
i_o(p)=\max\{L(p)-o,0\}.
\label{eq:app-token-offset}
\end{equation}
The position is determined separately for each prompt;
larger offsets select earlier positions.
For a CLIP sequence consisting of a beginning-of-text token,
content tokens, and an end-of-text token,
$o=1$ selects the end-of-text token and $o=2$ selects the last
content token.
Other tokenizers use the same index rule with their respective
non-padding sequence lengths.
Groups using pooled text representations are unaffected by $o$.

\paragraph{Confidence and Erased-Set Size Detection.}
Table~\ref{tab:fixed-hyperparameters} lists the fixed parameters
for confidence computation, erased-set size detection, and
numerical stabilization.
We use $\kappa=0.25$ for group weighting, self-calibration, and
size estimation, so $q_{1-\kappa}$ in Eq.~\eqref{eq:conf}
is the $75$th percentile.
The search uses a fixed horizon $J=96$.
For size detection, we set $B=8$ and $\alpha=0.30$:
the initial-rise condition requires increases in at least seven
of the first eight transitions, and the first sufficiently large
decline must reach $30\%$ of the largest log confidence drop.
The cutoff follows the rule in Section~\ref{sec:disclosure}.

\begin{table}[ht]
\centering
\small
\setlength{\tabcolsep}{5pt}
\caption{Fixed parameters for confidence computation,
erased-set size detection, and numerical stabilization
in the main configuration.}
\label{tab:fixed-hyperparameters}
\begin{tabularx}{\linewidth}{@{}clX@{}}
\toprule
Parameter & Value & Role \\
\midrule
$\kappa$ & $0.25$
& Specifies the upper-tail comparison range in the confidence statistic. \\
$J$ & $96$
& Sets the fixed search horizon. \\
$B$ & $8$
& Sets the window length for the initial-rise condition. \\
$\alpha$ & $0.30$
& Sets the relative threshold for the first sufficiently large decline. \\
\midrule
$\varepsilon_0$ & $10^{-3}$
& Sets the relative spectral cutoff in Eq.~\eqref{eq:fusion}. \\
$\varepsilon_c$ & $10^{-12}$
& Stabilizes the confidence denominator in Eq.~\eqref{eq:conf}. \\
$\varepsilon_z$ & $10^{-9}$
& Stabilizes gain standardization in Eq.~\eqref{eq:merge}. \\
$\varepsilon_g$ & $10^{-6}$
& Ensures positive group weights in Eq.~\eqref{eq:merge}. \\
$\varepsilon_a$ & $10^{-30}$
& Keeps logarithms defined in Eq.~\eqref{eq:confdrop}. \\
\bottomrule
\end{tabularx}
\end{table}

\paragraph{Datasets and Candidate Vocabularies.}
We construct \textbf{Imagenette-20} by extending the ten
Imagenette classes~\cite{imagenette} with ten non-overlapping
categories from ImageNet-1k~\cite{russakovsky2015imagenet}:
\emph{zebra}, \emph{giant panda}, \emph{pizza}, \emph{banana},
\emph{airliner}, \emph{school bus}, \emph{violin}, \emph{laptop},
\emph{submarine}, and \emph{umbrella}.
This expanded set enables evaluation across a wider range
of erased-set sizes.
Artistic-style identification covers the artists evaluated
by ESD~\cite{gandikota2023erasing} and UCE~\cite{gandikota2024unified}.

The erased concepts are drawn from these evaluation sets,
while identification searches a larger candidate vocabulary
$\mathcal C$.
For object recovery, the standard vocabulary contains the
1,000 ImageNet-1k class names~\cite{russakovsky2015imagenet}.
We construct an expanded vocabulary of 73,406 entries by
combining these class names with an English word list from
GNU Aspell\footnote{\url{https://aspell.net/}},
including its spelling-variant component.
This vocabulary covers all 1,000 ImageNet class names under
case-insensitive matching and retains multiword names and
capitalized proper names.
The 70,000-entry vocabulary used in the runtime comparison
is a subset of this expanded pool.
We will release the complete candidate vocabulary with our code.

\paragraph{Probe Selection and Prompt Templates.}
We use probe words for the representation-matching test
in Section~\ref{sec:assembly}.
Each probe is inserted into the applicable prompt templates
in $T$, and its representations are averaged across templates.
The test compares the inputs to edited matrices with the
corresponding text encodings to distinguish text-aware
from text-agnostic imprints.

Table~\ref{tab:prompt-templates} lists the template sets used
for probe matching and candidate encoding.
Each set contains five templates, including one that uses
the vocabulary entry without additional text.
Each placeholder is replaced with the complete entry.
Template averaging reduces sensitivity to phrasing,
as described in Section~\ref{sec:encoding}.
Artistic-style experiments are conducted separately with both
sets to assess sensitivity to the choice of templates.

\begin{table}[t]
\centering
\small
\setlength{\tabcolsep}{6pt}
\caption{Prompt templates for probe matching and candidate encoding.
Each placeholder is replaced with the complete vocabulary entry.}
\label{tab:prompt-templates}
\begin{tabular}{@{}cll@{}}
\toprule
Index & Object templates & Artistic-style templates \\
\midrule
1 & \texttt{a photo of a \{\}} & \texttt{art by \{\}} \\
2 & \texttt{\{\}} & \texttt{\{\}} \\
3 & \texttt{an image of a \{\}} & \texttt{painting by \{\}} \\
4 & \texttt{a picture of a \{\}} & \texttt{artwork by \{\}} \\
5 & \texttt{a photo of the \{\}} & \texttt{style of \{\}} \\
\bottomrule
\end{tabular}
\end{table}

\subsection{Extension to Multiword Concept Identification}
\label{app:multiword}

\textsc{Tracer} treats each entry in $\mathcal C$ as a complete
candidate, which may contain multiple words.
Identifying a multiword concept already present in $\mathcal C$
requires no change to the main pipeline.
This section describes an iterative search extension that
constructs candidate phrases from individual words.
The extension uses the text-aware footprint to guide expansion
and selects a phrase length without knowing the target length
in advance.

\paragraph{Candidate Initialization and Expansion.}
We construct a candidate word vocabulary $\mathcal V$
from alphabetic CLIP tokenizer entries ending with
\texttt{</w>}.
Each entry contains at least two characters after removing
the marker and is represented by a single token.
The initial candidate set $\mathcal C^{(1)}$ contains
all single-word entries in $\mathcal V$.
Candidates at expansion round $j$ contain exactly $j$ words.
This construction covers phrases whose constituent words
are each represented by a retained token.
At round $j$, we retain the $b$ highest-scoring candidates
in a set $\mathcal B^{(j)}$.
Each retained candidate is expanded by inserting a word
from $\mathcal V$ at every possible position:
\begin{equation}
\mathcal C^{(j+1)}
=
\left\{
\operatorname{Insert}(c,v,i)
\,\middle|\,
c\in\mathcal B^{(j)},\
v\in\mathcal V,\
i=0,\ldots,j
\right\}.
\label{eq:phrase-expansion}
\end{equation}
Here, $\operatorname{Insert}(c,v,i)$ inserts $v$ after
the first $i$ words of $c$, including before the first
word and after the last.
Duplicate sequences are removed, while distinct word
orders remain separate candidates.
Expansion draws from the full vocabulary $\mathcal V$,
allowing words with low individual scores to contribute
to a phrase.

\paragraph{Phrase Encoding and Footprint Scoring.}
At expansion round $j$, each candidate
$c=(v_1,\ldots,v_j)\in\mathcal C^{(j)}$ is encoded as
$[\mathrm{BOS},v_1,\ldots,v_j,\mathrm{EOS}]$,
where $\mathrm{BOS}$ and $\mathrm{EOS}$ mark the
beginning and end of the sequence.
We extract the CLIP hidden state at the final content-token
position, corresponding to $v_j$.
Causal attention allows this representation to incorporate
all preceding words.
The extension uses the bare phrase with a fixed extraction
position, without template averaging or token-offset search.
We center the candidate representations, apply regularized
covariance whitening with exponent $\gamma$, and perform
$\ell_2$ normalization.
The resulting unit vector for candidate $c$ at round $j$
is denoted by $\widetilde{x}_c^{(j)}$.
The mean and covariance are recomputed over
$\mathcal C^{(j)}$ at each expansion round.
The initial single-word representations can be reused
across checkpoints sharing the same base text encoder
and preprocessing settings.

Following Section~\ref{sec:assembly}, we fuse spectrally
reshaped weight footprints within the compatible text-aware
group to construct $G_{g_{\mathrm{txt}}}$.
This extension uses the $768$-dimensional CLIP input space.
Each candidate is scored by
\begin{equation}
s^{(j)}(c)
=
\bigl(\widetilde{x}_c^{(j)}\bigr)^\top
G_{g_{\mathrm{txt}}}
\widetilde{x}_c^{(j)}.
\label{eq:phrase-score}
\end{equation}
The quadratic form measures the footprint energy along
the candidate's normalized representation.
Dividing it by $\operatorname{tr}(G_{g_{\mathrm{txt}}})$
gives the singleton subspace coverage in this group.
We use the unnormalized score and standardize it within
each round as described below.

\paragraph{Cross-Length Comparison and Stopping.}
We standardize candidate scores within each expansion
round and apply a length penalty:
\begin{equation}
u^{(j)}(c)
=
\frac{s^{(j)}(c)-\mu_s^{(j)}}
{\sigma_s^{(j)}+\varepsilon_z}
-\lambda(j-1),
\label{eq:phrase-length-score}
\end{equation}
where $j$ denotes the current expansion round and the
number of words in each candidate.
The quantities $\mu_s^{(j)}$ and $\sigma_s^{(j)}$ are
the mean and standard deviation of scores over
$\mathcal C^{(j)}$, and $\varepsilon_z$ stabilizes
the denominator.
The standardized term measures a candidate's prominence
relative to other candidates of the same word length.
The coefficient $\lambda$ penalizes each additional word
beyond the first.
Within a round, this transformation preserves candidate
rankings and therefore leaves beam selection unchanged.

Let $c_j^\star=\arg\max_{c\in\mathcal C^{(j)}}u^{(j)}(c)$
denote the highest-scoring candidate at round $j$.
We initialize the best phrase with $c_1^\star$.
For each subsequent round $j\geq2$, we update the best
phrase only when
\begin{equation}
u^{(j)}(c_j^\star)
>
\max_{1\leq q<j}u^{(q)}(c_q^\star),
\label{eq:phrase-stopping}
\end{equation}
where $q$ indexes preceding expansion rounds and
$c_q^\star$ is the highest-scoring candidate at round $q$.
This condition requires the current round's best penalized
score to exceed the best score from all preceding rounds.
If it holds, we retain $c_j^\star$ and continue expansion
unless the maximum permitted length has been reached.
Otherwise, the search stops and returns the best phrase
from the preceding rounds.
We use a beam size of $b=8$, a length penalty of
$\lambda=0.05$, and a maximum of three words.

This stopping rule selects the word length of an individual
candidate.
It is separate from erased-set size detection, which estimates
the number of erased concepts.
Cross-length comparison is heuristic because both the
representation statistics and score distributions change
across rounds.
The search may therefore return a shorter phrase even
when the complete target has been generated.

\subsection{Baseline Implementations}
\label{app:base_impl}

We compare \textsc{Tracer} with brute-force generation
probing and an adapted membership inference attack~\citep{kong2024efficient}.
Both baselines assign a score to each candidate
$c\in\mathcal C$ based on changes between the public
base model $W$ and the unlearned model $W'$.
Candidates are ranked in descending order of these scores.

\paragraph{Brute-Force Generation Probing.}
This baseline evaluates whether a candidate concept becomes
less evident in generated images after unlearning.
For each candidate $c$, we construct prompts containing
the complete candidate name and generate images using
both $W$ and $W'$.
Let $\mathcal I_W(c)$ and $\mathcal I_{W'}(c)$ denote
the corresponding image sets.
We score the candidate by the decrease in average
concept evidence:
\begin{equation}
s_{\mathrm{BF}}(c)
=
\frac{1}{|\mathcal I_W(c)|}
\sum_{I\in\mathcal I_W(c)}a(I,c)
-
\frac{1}{|\mathcal I_{W'}(c)|}
\sum_{I\in\mathcal I_{W'}(c)}a(I,c),
\label{eq:baseline-bf}
\end{equation}
where $a(I,c)$ measures the evidence for concept $c$
in image $I$.
A larger score indicates a greater decline in the model's
ability to generate the candidate concept.
The same evaluation function is used for both models.

This procedure requires image generation for every
candidate in $\mathcal C$.
Changes in generated content may also reflect effects
on related non-target concepts, which can receive high
scores despite not being directly targeted.

\paragraph{Adapted Membership Inference Attack.}
We adapt PIA~\cite{kong2024efficient} from sample-level
membership inference to concept-level identification.
For each candidate $c$, let $\mathcal X_c$ denote
its associated evaluation samples.
We apply the PIA error statistic to the same samples
under $W$ and $W'$ and aggregate the differences:
\begin{equation}
s_{\mathrm{PIA}}(c)
=
\frac{1}{|\mathcal X_c|}
\sum_{x\in\mathcal X_c}
\left[
r_{W'}(x,c)-r_W(x,c)
\right],
\label{eq:baseline-pia}
\end{equation}
where $r_W(x,c)$ denotes the sample-level PIA error
computed using model $W$ with concept conditioning $c$.
A larger score indicates a greater increase in error
after unlearning.
Averaging over $\mathcal X_c$ converts the sample-level
changes into a single candidate score.

The adaptation ranks concepts by changes in the PIA
statistic rather than making binary membership decisions
for individual samples.
Its concept scores may capture both direct erasure
effects and changes affecting non-target concepts.

\newpage
\section{Additional Experimental Results}
\label{app:results}

Table~\ref{tab:k-t2i-art} reports the identification accuracy
of erased artistic styles across three text-to-image backbones.
\textsc{Tracer} achieves perfect identification in nearly
all evaluated settings.
On Stable Diffusion v1.5, ESD reaches $92\%$ when ten styles
are erased jointly, yielding an average accuracy of $98.0\%$.
All other evaluated configurations achieve $100\%$ accuracy.
UCE, ScaPre, and FIA maintain perfect identification
with twenty jointly erased styles.
These results demonstrate that weight footprints reveal
erasure targets beyond object concepts, extending
to artistic styles across different backbones and
unlearning methods.

\begin{table}[t]
\centering
\small
\setlength{\tabcolsep}{4pt}
\renewcommand{\arraystretch}{1.08}
\caption{Identification accuracy (\%) of erased artistic
styles across three text-to-image backbones.
$K$ denotes the number of jointly erased styles.
Shaded cells report evaluated results.
``--'' indicates an unevaluated setting.
Avg.\ is the arithmetic mean over evaluated settings
within each row.}
\label{tab:k-t2i-art}

\def\artvenue#1{{\scriptsize\sffamily #1}}
\def\eval#1{\cellcolor{gray!20}#1}

\begin{tabularx}{\linewidth}{@{}ll*{6}{>{\centering\arraybackslash}X}@{}}
\toprule
& & \multicolumn{5}{c}{Erased-set size $K$} & \\
\cmidrule(lr){3-7}
Method & Venue & 1 & 3 & 5 & 10 & 20 & Avg.$\,\uparrow$ \\
\midrule
\multicolumn{8}{c}{\textit{Stable Diffusion v1.5}} \\
\midrule
ESD
& \artvenue{ICCV~\citeyear{gandikota2023erasing}}
& \eval{100} & \eval{100} & \eval{100} & \eval{92} & -- & 98.0 \\
UCE
& \artvenue{WACV~\citeyear{gandikota2024unified}}
& \eval{100} & \eval{100} & \eval{100} & \eval{100} & \eval{100} & 100.0 \\
MACE
& \artvenue{CVPR~\citeyear{lu2024mace}}
& \eval{100} & \eval{100} & \eval{100} & \eval{100} & -- & 100.0 \\
FMN
& \artvenue{CVPR~\citeyear{zhang2024forget}}
& \eval{100} & \eval{100} & \eval{100} & -- & -- & 100.0 \\
ConceptPrune
& \artvenue{ICLR~\citeyear{chavhan2024conceptprune}}
& \eval{100} & \eval{100} & \eval{100} & \eval{100} & -- & 100.0 \\
Sculpting
& \artvenue{ICCV~\citeyear{li2025sculpting}}
& \eval{100} & \eval{100} & \eval{100} & \eval{100} & -- & 100.0 \\
SPEED
& \artvenue{ICLR~\citeyear{li2025speed}}
& \eval{100} & \eval{100} & -- & -- & -- & 100.0 \\
ScaPre
& \artvenue{ICLR~\citeyear{deng2026forget}}
& \eval{100} & \eval{100} & \eval{100} & \eval{100} & \eval{100} & 100.0 \\
FIA
& \artvenue{ICML~\citeyear{deng2026forget1}}
& \eval{100} & \eval{100} & \eval{100} & \eval{100} & \eval{100} & 100.0 \\
\midrule
AdvUnlearn
& \artvenue{NeurIPS~\citeyear{zhang2024defensive}}
& \eval{100} & -- & -- & -- & -- & 100.0 \\
STEREO
& \artvenue{CVPR~\citeyear{srivatsan2025stereo}}
& \eval{100} & \eval{100} & -- & -- & -- & 100.0 \\
Meta-Unlearning
& \artvenue{ICCV~\citeyear{gao2025meta}}
& \eval{100} & -- & -- & -- & -- & 100.0 \\
\midrule
\multicolumn{8}{c}{\textit{Stable Diffusion~3}} \\
\midrule
UCE
& \artvenue{WACV~\citeyear{gandikota2024unified}}
& \eval{100} & \eval{100} & \eval{100} & -- & -- & 100.0 \\
EraseFlow
& \artvenue{NeurIPS~\citeyear{kusumba2026eraseflow}}
& \eval{100} & \eval{100} & \eval{100} & -- & -- & 100.0 \\
\midrule
\multicolumn{8}{c}{\textit{FLUX.1}} \\
\midrule
EraseAnything
& \artvenue{ICML~\citeyear{gao2025eraseanything}}
& \eval{100} & \eval{100} & -- & -- & -- & 100.0 \\
ScaPre
& \artvenue{ICLR~\citeyear{deng2026forget}}
& \eval{100} & \eval{100} & \eval{100} & \eval{100} & -- & 100.0 \\
\bottomrule
\end{tabularx}
\end{table}

We exclude facial-identity unlearning from this evaluation.
In our preliminary checks, the tested methods suppress
target identities but also degrade the generation of
non-target content.
These checkpoints therefore provide a limited basis
for evaluating target identification under selective
unlearning, as their weight changes also reflect
broader losses in generative capability.



\newpage
\section{Code Sources for Unlearning and Attack Methods}
\label{app:code-sources}

We list the public code repositories for the unlearning
methods and the MIA baseline used in our experiments.

\begin{itemize}
    \setlength{\itemsep}{2pt}
    \setlength{\parskip}{0pt}

    \item ESD:
    \url{https://github.com/rohitgandikota/erasing}

    \item UCE:
    \url{https://github.com/rohitgandikota/unified-concept-editing}

    \item MACE:
    \url{https://github.com/Shilin-LU/MACE}

    \item FMN:
    \url{https://github.com/SHI-Labs/Forget-Me-Not}

    \item SPM:
    \url{https://github.com/THU-MIG/SPM}

    \item ConceptPrune:
    \url{https://github.com/ruchikachavhan/concept-prune}

    \item Sculpting:
    \url{https://github.com/coulsonlee/Sculpting-Memory-ICCV-2025}

    \item SPEED:
    \url{https://github.com/Ouxiang-Li/SPEED}

    \item ScaPre:
    \url{https://github.com/kaiyuan02415/ScaPre}

    \item FIA:
    \url{https://github.com/kaiyuan02415/Forget-It-All}

    \item AdvUnlearn:
    \url{https://github.com/OPTML-Group/AdvUnlearn}

    \item STEREO:
    \url{https://github.com/koushiksrivats/robust-concept-erasing}

    \item Meta-Unlearning:
    \url{https://github.com/sail-sg/Meta-Unlearning}

    \item EraseFlow:
    \url{https://github.com/Abhiramkns/EraseFlow}

    \item EraseAnything:
    \url{https://github.com/tomguluson92/EraseAnything}

    \item Refusal:
    \url{https://github.com/4idrossifenil-etanammide/Video-Unlearning}

    \item VideoEraser:
    \url{https://github.com/bluedream02/VideoEraser}

    \item SAFREE:
    \url{https://github.com/jaehong31/SAFREE}

    \item MIA baseline (PIA):
    \url{https://github.com/kong13661/PIA}
\end{itemize}

\end{document}